\documentclass[conference]{IEEEtran}
\IEEEoverridecommandlockouts
\usepackage{cite}
\usepackage{amsmath,amssymb,amsfonts}
\usepackage{algorithmic}
\usepackage{graphicx}
\usepackage{textcomp}
\usepackage{xcolor}
\usepackage{booktabs}
\usepackage{hyperref}
\def\BibTeX{{\rm B\kern-.05em{\sc i\kern-.025em b}\kern-.08em
    T\kern-.1667em\lower.7ex\hbox{E}\kern-.125emX}}
\begin{document}

\title{Multimodal Adaptive Retrieval Augmented Generation through Internal Representation Learning\\
}

\author{\IEEEauthorblockN{Ruoshuang Du}
\IEEEauthorblockA{\textit{Shanghaitech University} \\
\textit{School of Information Science and Technology}\\
dursh2024@shanghaitech.edu.cn}
\and
\IEEEauthorblockN{Xin Sun}
\IEEEauthorblockA{\textit{Chinese Academy of Sciences} \\
\textit{Institute of Automation}\\
sunxin000@mail.ustc.edu.cn}
\and
\IEEEauthorblockN{Qiang Liu}
\IEEEauthorblockA{\textit{Chinese Academy of Sciences} \\
\textit{Institute of Automation}\\
qiang.liu@nlpr.ia.ac.cn}
\and
\IEEEauthorblockN{Bowen Song}
\IEEEauthorblockA{\textit{Ant Group} \\
bowen.sbw@antgroup.com}
\and
\IEEEauthorblockN{Zhongqi Chen}
\IEEEauthorblockA{\textit{Ant Group} \\
chenzhongqi.czq@antgroup.com}
\and
\IEEEauthorblockN{Weiqiang Wang}
\IEEEauthorblockA{\textit{Ant Group} \\
weiqiang.wwq@antgroup.com}
\and
\IEEEauthorblockN{Liang Wang}
\IEEEauthorblockA{\textit{Chinese Academy of Sciences} \\
\textit{Institute of Automation}\\
wangliang@nlpr.ia.ac.cn}
}

\maketitle

\begin{abstract}
Visual Question Answering systems face reliability issues due to hallucinations, where models generate answers misaligned with visual input or factual knowledge. 
While Retrieval Augmented Generation frameworks mitigate this issue by incorporating external knowledge, static retrieval often introduces irrelevant or conflicting content, particularly in visual RAG settings where visually similar but semantically incorrect evidence may be retrieved. 
To address this, we propose Multimodal Adaptive RAG (MMA-RAG), which dynamically assesses the confidence in the internal knowledge of the model to decide whether to incorporate the retrieved external information into the generation process. 
Central to MMA-RAG is a decision classifier trained through a layer-wise analysis, which leverages joint internal visual and textual representations to guide the use of reverse image retrieval. 
Experiments demonstrated that the model achieves a significant improvement in response performance in three VQA datasets.
Meanwhile, ablation studies highlighted the importance of internal representations in adaptive retrieval decisions.
In general, the experimental results demonstrated that MMA-RAG effectively balances external knowledge utilization and inference robustness in diverse multimodal scenarios.
We make all code and data publicly available at \href{https://anonymous.4open.science/r/Multimodal-Adaptive-RAG-20AB/}{github}.

\end{abstract}


\section{Introduction}
\label{sec:intro}
Large language models (LLMs) have achieved remarkable success in a wide range of natural language understanding and generation tasks \cite{shao2024survey}.  
However, despite their impressive capabilities, LLMs are known to suffer from a critical limitation: hallucination \cite{li2025him, huang2025survey, park2024mitigating}. 
This phenomenon refers to the generation of outputs that are factually inaccurate, unverifiable, or inconsistent with the provided input. 

To address this limitation, Retrieval-Augmented Generation (RAG) has emerged as a promising solution \cite{huang2024survey, gao2023retrieval}. 
RAG improves language model performance by incorporating external knowledge retrieved from large-scale textual corpora to complement the model's internal parameterized representations \cite{lewis2020retrieval, wu2024retrieval, 11227249}. 
This approach has been shown to improve response accuracy and factual consistency, particularly in knowledge-intensive tasks where reliance on static, pre-trained parameters alone may lead to outdated or incorrect information \cite{zhao2024retrieval}. 

Although early implementations were confined to purely textual settings, recent work has extended this idea to multimodal contexts, leading to multimodal RAG frameworks \cite{zhao2023retrieving, yasunaga2022retrieval, yu2025mramg, hu2023reveal, zhai2025sam}. 
In multimodal RAG, models generate image-question conditioned retrieval queries, sourcing both text and images from external databases to support reasoning. 
These systems have shown promise on complex tasks such as Visual Question Answering (VQA) \cite{antol2015vqa}. 
For example, REVIVE uses regional visual representations combined with the knowledge retrieved to improve the accuracy of the answers \cite{lin2022revive}. 
Similarly, MuRAG improves response generation by incorporating information from external knowledge bases, enabling more effective joint reasoning over images and texts \cite{chen2022murag}. 
The multimodal alignment model introduced a multimodal large language model-based reclassification step, selecting the most relevant knowledge from the top candidates retrieved to improve performance \cite{chen2025seeing}. 

Reverse Image Retrieval (RIR) can be viewed as a specialized variant of multimodal RAG, in which additional multimodal context is provided by retrieving visually similar images from web-based sources \cite{xu2024reverse}. 
In this approach, screenshots or related images retrieved based on the query image are integrated with the original input, thereby enriching the model's input space with complementary visual information. 
Importantly, the advantage of RIR does not typically lie in directly supplying the correct answer, but rather in facilitating more accurate grounding and interpretation of the query through contextually relevant visual cues. 

\begin{figure*}[t]
  \centering
  \includegraphics[width=\linewidth]{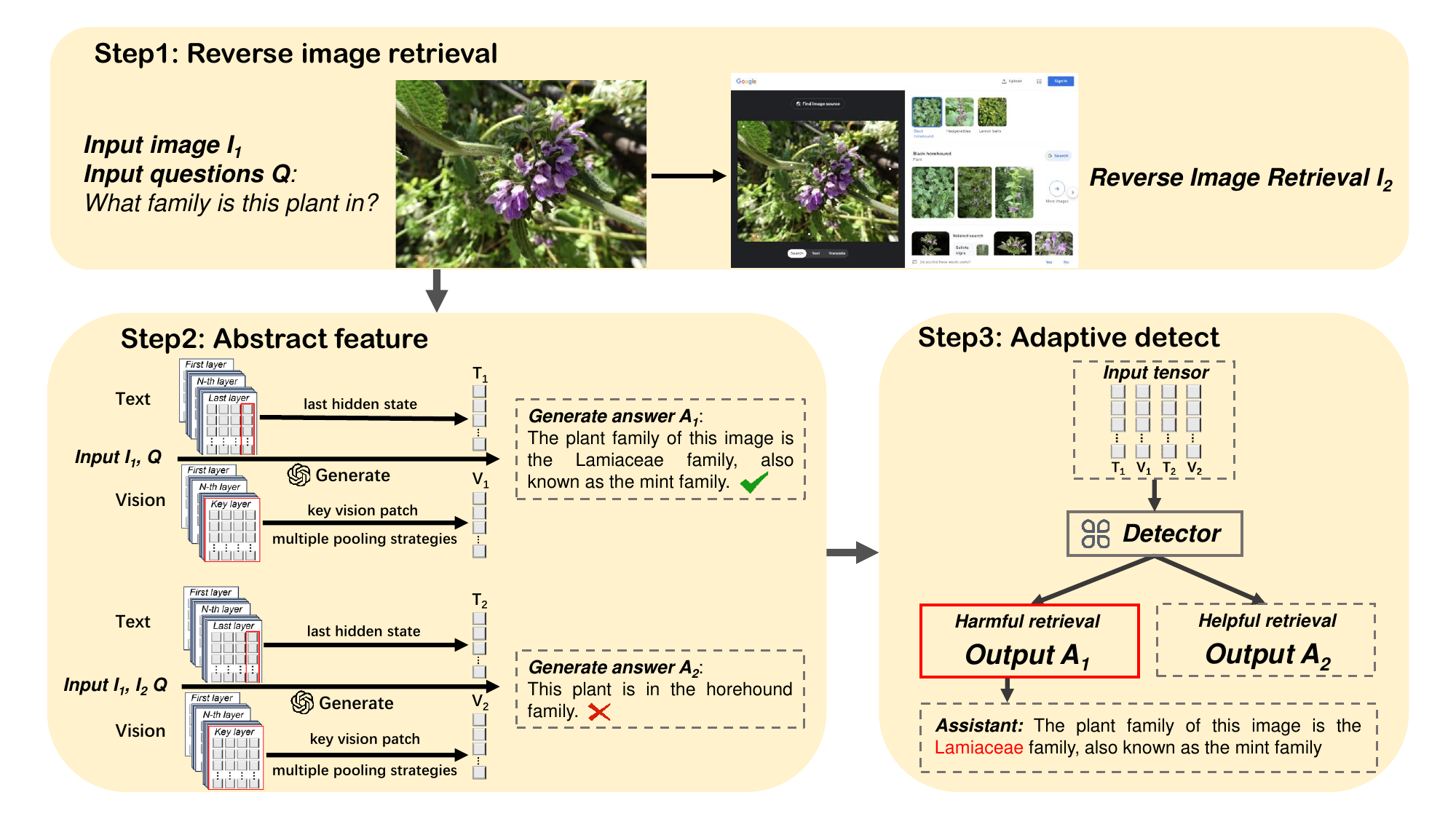}
  \caption{Overview of the Multimodal Adaptive RAG framework}
  \label{main-figure}
\end{figure*}

Despite the aforementioned advantages, visual retrieval augmented generation introduces a more challenging failure mode than its text-only counterpart.
Images returned by visual retrieval are often highly similar in appearance, yet semantically inconsistent with the query.
As illustrated in Fig.~\ref{main-figure}, queries concerning plants from the \textit{Lamiaceae} family may retrieve visually similar species such as Horehound, resulting in evidence that appears highly convincing but is in fact incorrect.
This phenomenon, referred to as \emph{visual similarity with semantic mismatch}, is more difficult to defend against than interference in text-based RAG systems.
Consequently, effective mitigation requires joint reasoning over visual and textual features, enabling the model to simultaneously assess the visual similarity of retrieved content and its semantic consistency with the query.

Similar to RIR, many existing multimodal RAG methods implicitly assume that external information is always beneficial, often introducing irrelevant or misleading evidence—particularly in cases where the model already possesses sufficient internal knowledge—thereby causing retrieval redundancy that is likely to degrade overall model performance \cite{li2024benchmarking, jiang2023active, mei2025survey}.

In this paper, we propose Multimodal Adaptive Retrieval Augmented Generation (MMA-RAG) designed to address the challenges of avoiding harmful factors caused by visually similar but semantically incorrect evidence and the rational use of retrieved external information. 
Specifically, MMA-RAG first extracts the hidden states of textual and visual features, aligns them, and integrates them into a unified vision–language joint representation.
Based on this representation, we train a four-class classifier to predict the impact of retrieval on correctness of the answer.
Crucially, we argue that simple last-layer representations may not fully capture the nuanced misalignment between visual and textual modalities.
Through a comprehensive layer-wise analysis of the model's internal states, we observe that the semantic alignment between vision and text evolves differently across network depths.
Text-only features show limited discriminative capability in shallow layers and become effective only in deeper layers, whereas multimodal features achieve high detection accuracy even in the early layers.
This observation indicates that multimodal fusion is crucial for the early identification of erroneous or misleading evidence.
Motivated by this finding, MMA-RAG strategically selects and fuses the most informative intermediate representations to train a multimodal joint-feature classifier.
Finally, guided by the classifier’s predictions, the model adaptively employs the reverse image retrieval mechanism, avoiding the introduction of harmful external information while ensuring the generation of correct answers whenever possible.
The core contributions of this paper are summarized as follows:
\begin{itemize}
\item We propose MMA-RAG, a multimodal adaptive retrieval augmented generation framework that predicts the utility of RIR from internal multimodal representations to mitigate harmful retrieval in visual question answering tasks.

\item We perform a layer-wise analysis of multimodal large language models, revealing how visual and textual confidence signals evolve and informing the selection of internal features for hallucination detection.

\item We design an internal-representation-based retrieval utility classifier that integrates multimodal features to assess whether external retrieval improves response correctness.

\item Extensive experiments on three knowledge-intensive VQA benchmarks with multiple vision--language backbone models demonstrate that MMA-RAG outperforms standard retrieval-based methods and existing baselines.

\end{itemize}

\section{Methodology}
\label{sec:Methodology}

Although RIR can provide valuable external context by retrieving visually similar images, it can also introduce harmful samples that mislead the model, especially when the retrieved images contain semantically irrelevant or contradictory content. 
In such cases, relying on external retrieval may cause the model to generate incorrect answers, even when the original image alone would have sufficed for correct reasoning. 
As illustrated in the example depicted in Fig.~\ref{main-figure}, the use of RIR yields the answer "The plant is in the horehound family", which is in fact incorrect. 
The accurate response corresponds to the case without RIR, namely, "The plant family of this image is the Lamiaceae family, also known as the mint family".  

To address such issues in VQA tasks, we propose a \textbf{Multi-modal Adaptive Retrieval-Augmented Generation (MMA-RAG)} framework to address the challenge of harmful retrievals in VQA tasks. 
The core idea is to adaptively determine whether externally retrieved images should be incorporated into the generation process to minimize the negative impact of irrelevant or misleading visual information. 
If the retrieved images are helpful to improve the accuracy of the answers, the MMA-RAG framework adopts the RAG approach, incorporating all images and the corresponding question as input to the generation model. 
If the retrieved images introduce noise and degrade answer quality, the framework relies solely on the original image and the question for answer generation. 

MMA-RAG is designed to make adaptive use of retrieved visual content in a multimodal VQA setting. 
It consists of three key components: 

\textbf{1) Reverse Image Retrieval}: 
For each VQA instance, the input consists of a query $Q$ and an image $I_1$. 
We perform reverse image retrieval using $I_1$ by querying visually similar images on Google and capturing screenshots of the retrieved results. 
These screenshots serve as an additional input image $I_2$, which may subsequently be fed into the large model along with the original question $Q$ and image $I_1$. 

\begin{figure}[h]
  \centering
  \includegraphics[width=\linewidth, , trim=0 40 0 20, clip]{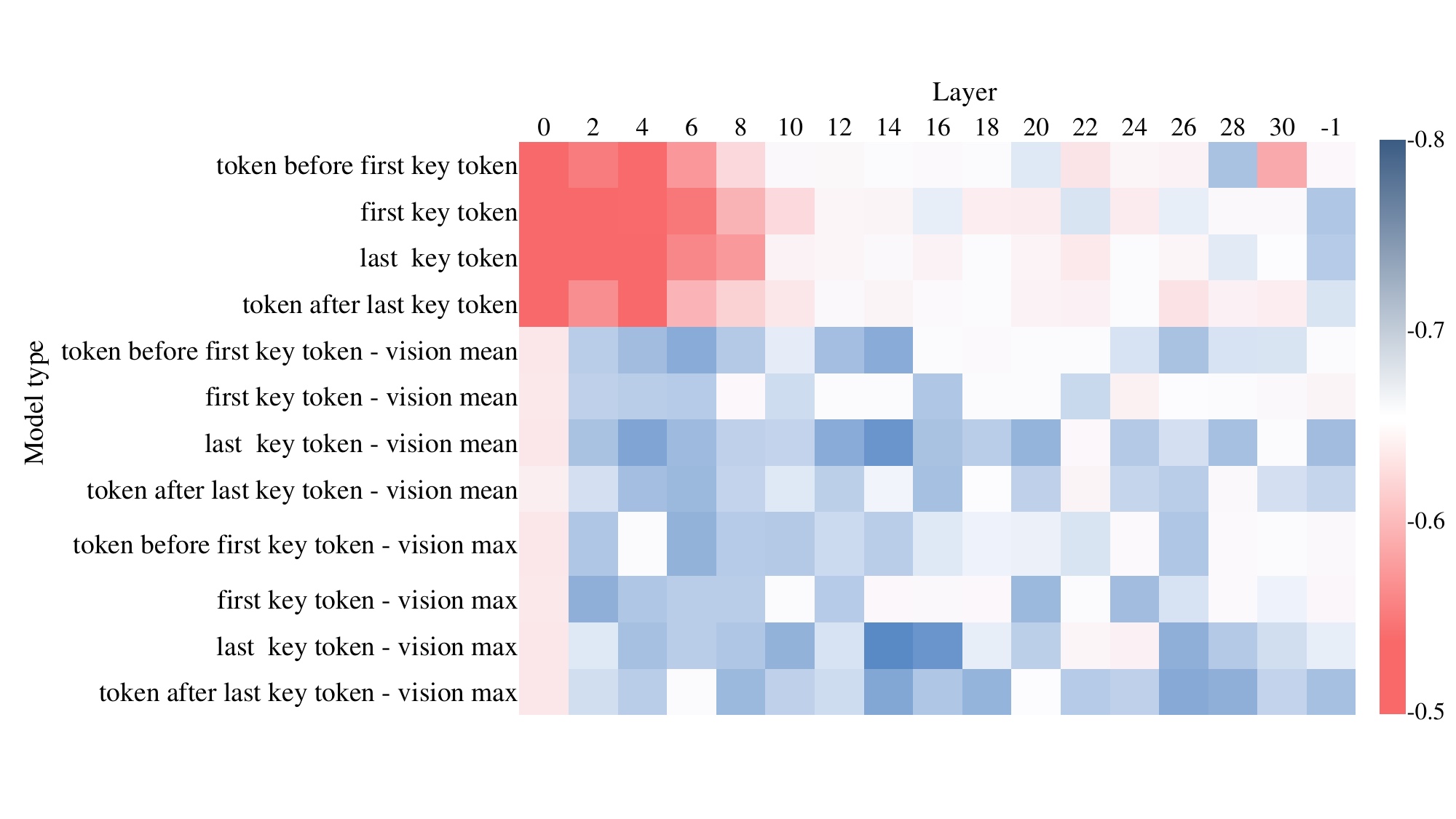}
  \caption{Heatmap of the classifier accuracy across different key tokens and layers on the OK-VQA dataset using the Idefics2-8B model}
  \label{fig:heatmap}
\end{figure}

\textbf{2) Abstract feature}: 
In purely text-based large language models, the hidden states of the exact answer token can serve as key tokens for error detection\cite{orgad2024llms}.
We extend this observation to multimodal large language models and perform a layer-wise analysis, showing that multimodal fusion leads to more accurate error detection in multimodal settings.
Specifically, we evaluate error-detection capability across different layers using the Idefics2-8B backbone on the OK-VQA dataset. Hidden states are extracted from every even-numbered Transformer layer as well as the final layer.
We compare multiple feature configurations: textual hidden states at key token positions alone, and textual states combined with vision features using different pooling methods.
As shown in Fig.~\ref{fig:heatmap}, the results reveal several key insights into the internal decision-making process of the model. 

\textbf{Primacy of Multimodal Fusion.}
Error-detection methods that integrate visual features consistently outperform those based solely on textual hidden states across all layers.
This observation indicates that visual representations play a critical role in assessing the internal certainty of the model and in deciding whether external retrieval is required.

\textbf{Layer-wise Evolution of Information.}
Text-only features exhibit limited predictive capability in shallow layers ranging from layer 0 to layer 10, but their performance improves substantially in deeper layers. In contrast, multimodal features reach high accuracy much earlier, particularly within the intermediate layers from layer 2 to layer 16. This trend suggests that the alignment between visual and textual cues becomes sufficiently established at mid-network depths to effectively support retrieval gating.

\textbf{Limited Sensitivity to Pooling Strategies.} Both average and maximum pooling over visual features result in similar high-accuracy regions for error detection. This suggests that, when visual information is globally aggregated, the detector mainly relies on image-level semantics, while the specific pooling operator has only a minor impact.
Therefore, we leverage the model's internal textual and visual hidden states to train the adaptive classifier, as these representations preserve rich semantic and contextual information.
Although key tokens serve as effective diagnostic signals that represent posterior knowledge in retrospective analysis, they are inaccessible during the inference phase of an adaptive system. Consequently, for the textual feature $T_1$, we utilize the hidden state from the final decoding step.
This state synthesizes information from the question, the image, and the generated prefix, thereby naturally reflecting the model's current belief state.
For the image feature, after feeding the input image into the model, multiple layers of visual representations are obtained. 
We select a specific intermediate layer and compute the average pooling over all patch embeddings within that layer. 
The resulting vector serves as the compact and informative vision feature $V_1$ that represents the input image.
We jointly input $Q$, $I_1$, and $I_2$ into the model, following a similar procedure as described above, to extract the corresponding textual characteristic $T_2$ and vision feature $V_2$. 

To effectively capture both semantic and visual cues, we concatenate the generated feature $T_1$, $V_1$, $T_2$, $V_2$ to form a unified representation $H_c$ for classification. 
$$
{H}_{c} = \text{Concat}(T_1, V_1, T_2, V_2)
$$

\textbf{3) Adaptive detect}: 
The extracted data $H_c$ are utilized to train a four-class classifier. 
Considering both performance and efficiency, we adopt a multilayer perceptron architecture for the classifier, which is employed to evaluate the retrieval utility under different conditions as a retrieval gating mechanism. 
$$
\hat{y}=\arg \max _y \operatorname{Softmax}(f(H_c))
$$
where $\hat{y}$ represents the classification result. 
Specifically, there are four possible scenarios:
1) $S_1$: Both using external retrieval and not using external retrieval result in an incorrect answer.
2) $S_2$: Using external retrieval leads to a correct answer, while not using it results in an incorrect answer.
3) $S_3$: Using external retrieval results in an incorrect answer, while not using it leads to a correct answer.
4) $S_4$: Both using external retrieval and not using external retrieval result in the correct answer.

At inference time, we introduce two alternative retrieval trigger strategies based on the classifier’s four-way prediction. 
These strategies represent two opposing preferences regarding the reliance on external retrieval. 

\textbf{RIR-Pessimistic Strategy. }
This pessimistic-oriented strategy adopts a cautious stance toward external retrieval. 
Retrieval is triggered only when it is predicted to be essential, i.e. when using retrieved images leads to a correct answer, and not using them would result in an incorrect one. 
In all other scenarios, the model discards the retrieved content and relies solely on the original image and the question. 
This strategy minimizes the risk of introducing harmful noise, favoring the default path unless a clear benefit is identified. 
$$
R= \begin{cases}1 & \text { if }  \hat{y} =S_2 \\ 0 & \text { if  } \hat{y}\neq S_2 \end{cases}
$$
where R decides whether to use RIR. 

\textbf{RIR-Optimistic Strategy. }
This optimistic oriented strategy takes a more liberal approach to retrieval usage, reflecting a retrieval-favoring bias. 
It triggers external retrieval in all cases except when it is predicted to degrade performance, that is, when the retrieved images introduce noise and hurt the response quality. 
In this view, an external visual context is generally helpful and should be included unless explicitly harmful. 
$$
R= \begin{cases}1 & \text { if }  \hat{y} \neq S_3 \\ 0 & \text { if  } \hat{y}= S_3 \end{cases}
$$
When R is 1, choose to use RIR to generate the answer, that is, input the original image, screenshot image and question together into the model to generate the answer; otherwise, do not use the screenshot image. 

The two strategies embody different stances toward external retrieval.
Our subsequent experiments investigate their hierarchical impact across different datasets.
This flexible decision layer enables the system to trade off robustness and completeness while adapting to the specific characteristics of the data or application domain.

\section{Experiments}
\label{sec: Experiments}

In this section, we evaluate the effectiveness of the multimodal adaptive classifier across multiple datasets. 
We use model-judged accuracy as metrics for answer correctness. 
We systematically compare the performance of four configurations: zero-shot prompt, few-shot prompt, few-shot prompt with RIR, and our proposed method MMA-RAG, which adaptively determines whether to apply RIR based on a classifier under the few-shot setting.

\subsection{Datasets and Knowledge Bases}

Visual question answering tasks that require external knowledge integration face unique challenges, as answers depend on information beyond image content. 
Foundational benchmarks such as OK-VQA \cite{marino2019ok} exemplify this paradigm, featuring more than 14000 questions that demand reasoning with common sense and domain-specific knowledge. 
In this dataset, there have already been many excellent studies. 
Methods for extracting relevant knowledge from noisy sources \cite{wu2022multi}, cross-modal retrieval frameworks \cite{chen2022murag}, and hybrid graph representations \cite{zhu2020mucko} demonstrate improved performance in knowledge-intensive VQA tasks, highlighting the necessity of unified visual-textual knowledge architectures. 

To address the limitations of existing datasets in the evaluation of deep visual knowledge integration, Encyclopedic-VQA introduces a comprehensive benchmark featuring 221K unique question-answer pairs, each linked to up to 5 images \cite{mensink2023encyclopedic}. 
These images are derived from iNaturalist 2021 \cite{van2021benchmarking} and Google Landmarks Dataset V2 \cite{weyand2020google}. 
Unlike previous VQA tasks that focus on understanding a generic scene, this dataset emphasizes fine-grained instance-level attributes that require encyclopedic knowledge of specific entities, artifacts, or biological species. 
Encyclopedic-VQA has become a standard testbed for evaluating knowledge-intensive vision systems. 
Recent works like EchoSight \cite{yan2024echosight} achieve a more effective fusion of multimodal knowledge, validating the practicality of the Encyclopedic VQA data set to advance the synthesis of multimodal knowledge. 

InfoSeek \cite{chen2023can} establishes a large-scale benchmark for visual information search queries, consisting of approximately 1.3 million questions grounded in over 11000 visual entities derived from the Open Visual Entity Nexus dataset \cite{hu2023open}. 
These entities span diverse domains, including cultural landmarks, rare species, and technological artifacts, ensuring broad coverage of knowledge-intensive topics. 
The dataset combines 8.9K human-annotated QA pairs with 1.3M automatically generated questions, using templated parsing of Wikipedia infoboxes and entity-attribute relationships. 

Our dataset is randomly sampled from the aforementioned public datasets, followed by obtaining screenshots to prepare the data for training and evaluation. 
The numbers of training and evaluation samples for the three datasets are summarized in Table 1.

\begin{table}
  \centering
  \caption{Training and evaluation sample sizes of the three datasets.}
  \begin{tabular}{cccc}
    \toprule
    Dataset& Infoseek& OK-VQA& E-VQA\\
    \midrule
    Training& 3740& 1000& 1000\\
    Evaluation& 1646& 4989& 3345\\
    \bottomrule
  \end{tabular}
  \label{tab:Training and evaluation sample sizes of the three datasets}
\end{table}

\subsection{Backbone Models and Metrics}

Idefics2-8B is an open source vision language model introduced by Hugging Face, featuring 8 billion parameters \cite{laurenccon2024matters}. 
This model can process arbitrary sequences of text and image input to generate textual outputs. 
In multiple visual question answering benchmarks, Idefics-2 ranks among the top models of similar scale, with performance comparable to larger models. 

Idefics3-8B is a vision language model developed by the research team at Hugging Face, designed to process image and text input while generating textual output \cite{laurenccon2024building}. 
Architecturally, Idefics3-8B uses a cross-attention mechanism, which allows an effective integration of visual and linguistic information, thus achieving strong performance in multimodal tasks. 

The Qwen2-VL series introduces several key advancements in vision language models to enhance the model’s ability to process images of varying resolutions \cite{wang2024qwen2}. 
Qwen2.5-VL extends these capabilities by introducing a unified vision framework for both images and videos, multimodal rotary position embedding (M-RoPE) for improved cross-modal alignment, and a Naive Dynamic Resolution mechanism for adaptive visual tokenization based on image resolution. 

We employ Qwen2.5-Instruct as an automatic evaluator to assess the correctness of the generated responses, based on which we compute the accuracy for each dataset.

\subsection{Baselines}
\textbf{RIR} generates responses by incorporating reverse image retrieval \cite{xu2024reverse}.
Building upon RIR, we further consider several baselines that make adaptive decisions about whether to use RIR.
\textbf{Chain-of-Thought (CoT)} encourages the model to produce explicit intermediate reasoning steps \cite{wei2022chain}.
\textbf{P(true)} serves as a standard confidence-based baseline, which estimates correctness by calculating the probability that the model affirms its own answer \cite{kadavath2022language}.
\textbf{CLIP} measures image–text semantic alignment using joint embeddings \cite{radford2021learning}.

\section{Results}
\label{sec:Results}

\subsection{Main Results}
We trained the classifier for Multimodal Adaptive Retrieval Augmented Generation using visual features and textual hidden states and conducted experiments on three datasets: InfoSeek, OK-VQA, and Encyclopedic-VQA. 
Given the varying class distributions across datasets, we apply dataset-specific class weights to balance the contribution of different classes during training, thereby ensuring the reliability and validity of the experiments. 
The retrieval process involved obtaining similar images through Google search, capturing screenshots and using these screenshots alongside the original images and questions as input to generate final answers. 
The classifier's role was to assess the utility of such retrievals; if beneficial for generating correct answers, external retrieval was employed; otherwise, it was omitted. 
During the testing process, Qwen2.5-instruct was used to evaluate whether the answers were correct and the final evaluation metric was accuracy. 
Tab.~\ref{tab:MM_adaptive_RAG} shows the main experimental results. 
\begin{table}
  \centering
  \caption{Accuracy scores on the InfoSeek, E-VQA, and OK-VQA test datasets, using Idefics2-8B, Idefics3-8B, and Qwen2VL-7B as backbones, with Qwen25\_Instruct used to judge answer correctness. \textbf{Bold} indicates state-of-the-art performance} 
  \begin{tabular}{ccccc}
    \toprule
    Model & Method & InfoSeek & OK-VQA & E-VQA \\
    \midrule
    & Zero shot& 19.9& 58.7
& 10.0\\
 & Few shot& 15.9& 58.5&14.1\\
 & RIR& 23.3 & 62.2 &19.8 
\\
 Qwen2VL& CoT& 20.4& 46.7&14.4\\
 & P(true)& 22.6& 53.3&19.1\\
 & CLIP& 20.9& 56.0&18.5\\
    & \textbf{MMA-RAG}& \textbf{23.9} & \textbf{62.4} & \textbf{20.0} \\
    \midrule
    & Zero shot
& 15.1& 53.8& 13.0\\
 & Few shot
& 14.2& 58.7&12.7\\
 & RIR
& 17.2& 56.7&14.1\\
 Idefics2& CoT
& 15.5& 49.0&14.1\\
 & P(true)
& 14.1& 49.1&13.8\\
 & CLIP& 16.4& 54.8&14.3\\
    & \textbf{MMA-RAG}& \textbf{20.3}& \textbf{60.1}& \textbf{14.6}\\
    \midrule
    & Zero shot
& 12.5& 43.2& 11.9\\
 & Few shot
& 8.5& 46.0&6.3\\
 & RIR
& 17.5 & 56.6 &12.5 
\\
 Idefics3& CoT
& 14.3& 43.7&10.9\\
 & P(true)
& 12.9& 43.7&11.0\\
 & CLIP& 12.6& 50.0&10.0\\
    & \textbf{MMA-RAG}& \textbf{18.1} & \textbf{58.3} & \textbf{12.6} \\
        \bottomrule
  \end{tabular}
  \label{tab:MM_adaptive_RAG}
\end{table}

Across different models and datasets, the performance of zero-shot and few-shot prompts is inconsistent, primarily due to variations in dataset types and task characteristics. 
To ensure experimental validity, we adopted the few-shot prompt in both the RIR and MMA-RAG experiments.
Reverse Image Retrieval systems that use Google to retrieve similar images have significantly improved the accuracy of the generated answers. 
However, in certain cases, the use of retrieved external images leads to a degradation in response quality, producing incorrect responses that would not have occurred if the model had relied solely on the original image and question. 
We refer to such instances as "harmful samples", where the introduction of additional visual information inadvertently misguides the model. 
Compared with the original Reverse Image Retrieval model, the multimodal adaptive RAG model achieves varying degrees of improvement in three datasets. 
Compared with the confidence-based P(true) baseline, the auxiliary-model-based CLIP approach, and reasoning-based CoT method, MMA-RAG consistently delivers substantial and statistically significant performance gains across all evaluated datasets.
This improvement is primarily attributed to the model’s ability to predict and suppress the influence of harmful samples, thus avoiding unnecessary or detrimental retrievals and preserving the model's ability to generate correct answers.

For certain model configurations on OK-VQA and E-VQA, MMA-RAG yields only marginal improvements, likely because the proportion of harmful samples introduced by RIR is low, leaving limited room for denoising.
This also indicates that MMA-RAG is relatively safe: when retrieval is beneficial, the gating mechanism tends to keep retrieval enabled and is unlikely to introduce adverse effects.

\subsection{Feature Robustness}

We further conducted exploratory experiments on the extraction of visual features, including selecting patch embeddings from different Transformer layers and applying various pooling strategies, such as average pooling and max pooling to obtain global image representations. 

\begin{figure}[h]
  \centering
  \includegraphics[width=\linewidth]{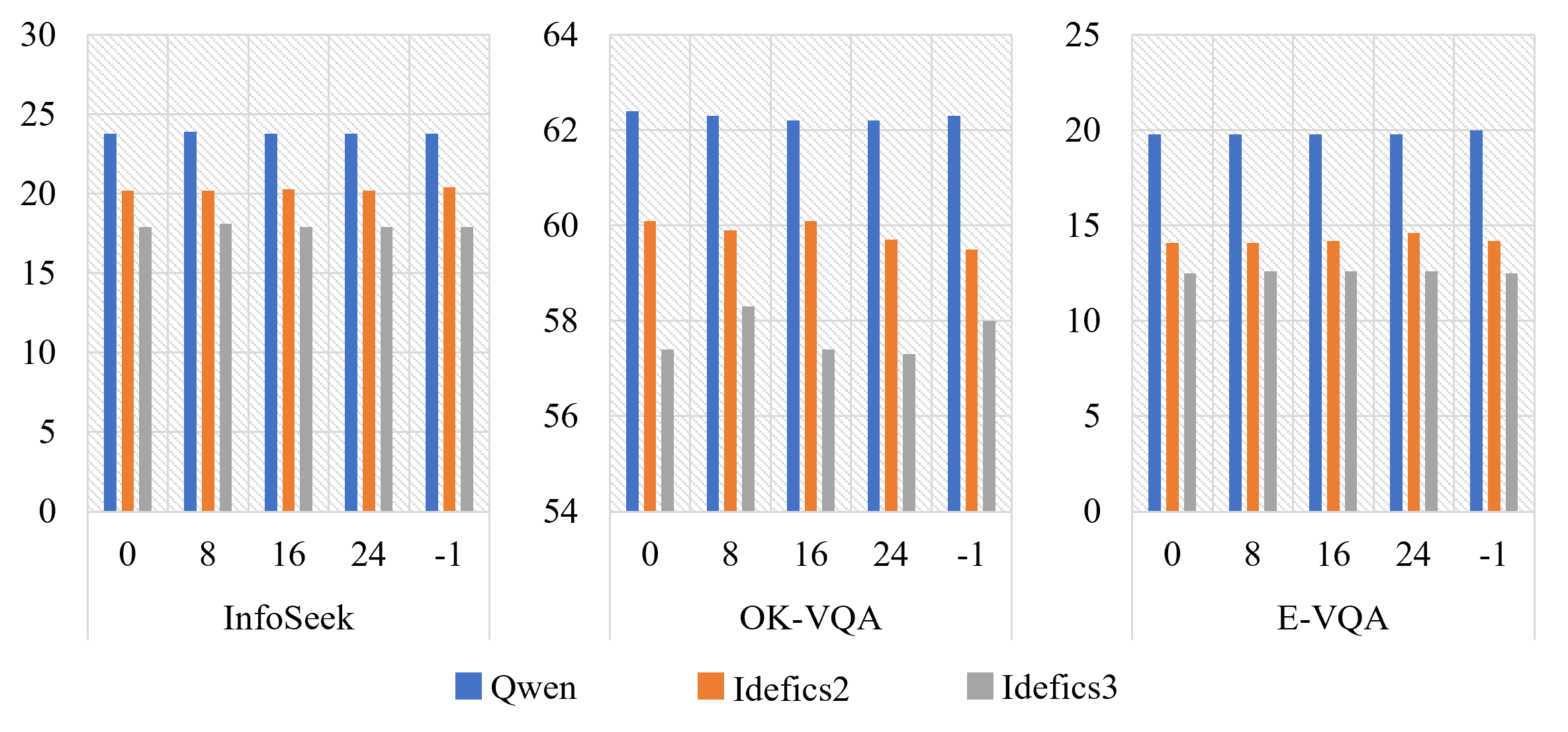}
  \caption{Impact of Transformer Layer Selection on Visual Feature Representation }
  \label{column-figure}
\end{figure}

As illustrated in Fig.~\ref{column-figure}, the vertical axis represents the response accuracy and the horizontal axis corresponds to the layer index.
Despite differences in feature extraction strategies, overall performance variation remained relatively small. 
This limited variance can be attributed to the robustness of the downstream classifier to feature extraction strategies. 
Since the middle and later layers of vision transformers already produce stable semantic representations, and pooling differences are largely smoothed out by the subsequent MLP, the overall performance is less sensitive to such variations.

\subsection{Ablation Study}

The consistent improvements observed across the datasets can be attributed to the effective exploitation of internal hidden states by the multimodal adaptive RAG model. 
In particular, both the textual hidden states, which encapsulate the semantic representation and contextual reasoning derived from the question and the generated response, and the visual features, which encode salient spatial and perceptual information from the input image, play a pivotal role in informing adaptive retrieval decisions. 
To better understand the individual and combined contributions of these features, we conducted a series of ablation studies, analyzing their respective impacts on the classifier's performance within the MMA-RAG framework. 

In this experiment, we trained the classifier on three datasets using only hidden textual states or only visual features, while keeping all other settings unchanged. 
\begin{table}
  \centering
  \caption{Performance of ablation study on Multimodal adaptive RAG with idefics2-8B. The performance of the model is jointly influenced by textual and visual features.}
  \begin{tabular}{cccc}
    \toprule
    Method & Infoseek& OK-VQA& E-VQA\\
    \midrule
    RIR & 17.2 & 56.7 & 14.1 \\
    wo-text & 17.9 & 58.6 & 14.2 \\
    wo-vision & 19.3 & 59.9 & 14.2 \\
    \textbf{MM Adaptive RAG}& \textbf{20.3} & \textbf{60.1} & \textbf{14.6} \\
    \bottomrule
  \end{tabular}
  \label{tab:MM_adaptive_RAG_ablation_study}
\end{table}

As shown in the results of Tab.~\ref{tab:MM_adaptive_RAG_ablation_study} , given the limited accuracy of large models in generating answers, classifiers that incorporate visual features outperform those relying solely on hidden text states. 
This suggests that in VQA tasks, extracting and utilizing visual features may enhance the ability to determine the effectiveness of external retrievals. 
Similarly, classifiers that lack textual hidden states have also exhibited a decline in performance. 
This has demonstrated that hidden textual states inherently contain implicit cues regarding the accuracy of the generated answers. 
Therefore, by effectively integrating hidden text states with visual features, we were able to extract maximally valuable information, thus enhancing our understanding of the model decision-making process. 

\begin{figure}[h]
  \centering
  \includegraphics[width=\linewidth, trim=0 20 0 20, clip]{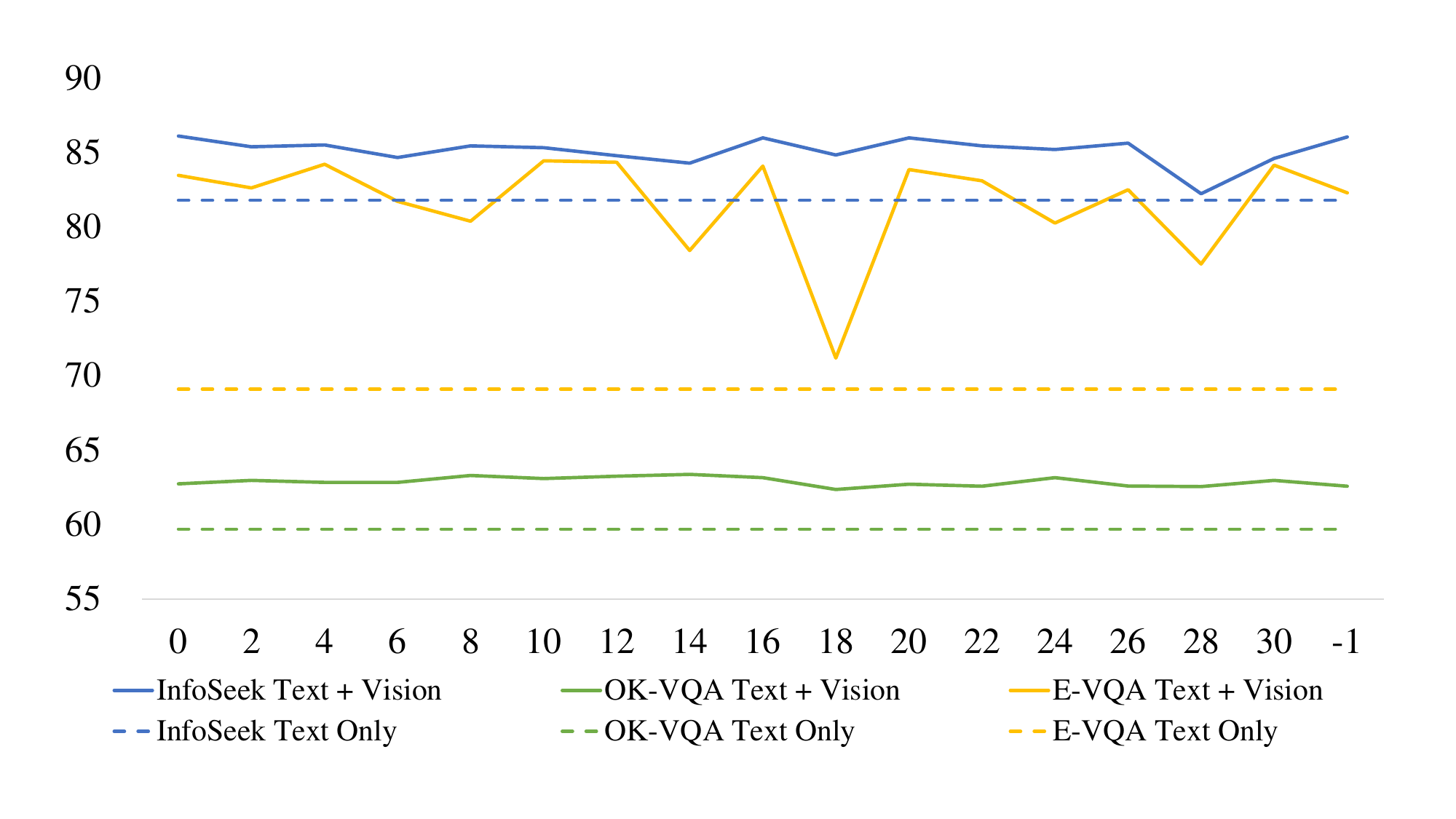}
  \caption{Layer-wise Comparison of Classifier Accuracy Using Textual and Multimodal Features on the InfoSeek, OK-VQA, and E-VQA Datasets}
  \label{classifer}
\end{figure}

\begin{figure}[h]
  \centering
  \includegraphics[width=\linewidth, trim=0 20 0 20, clip]{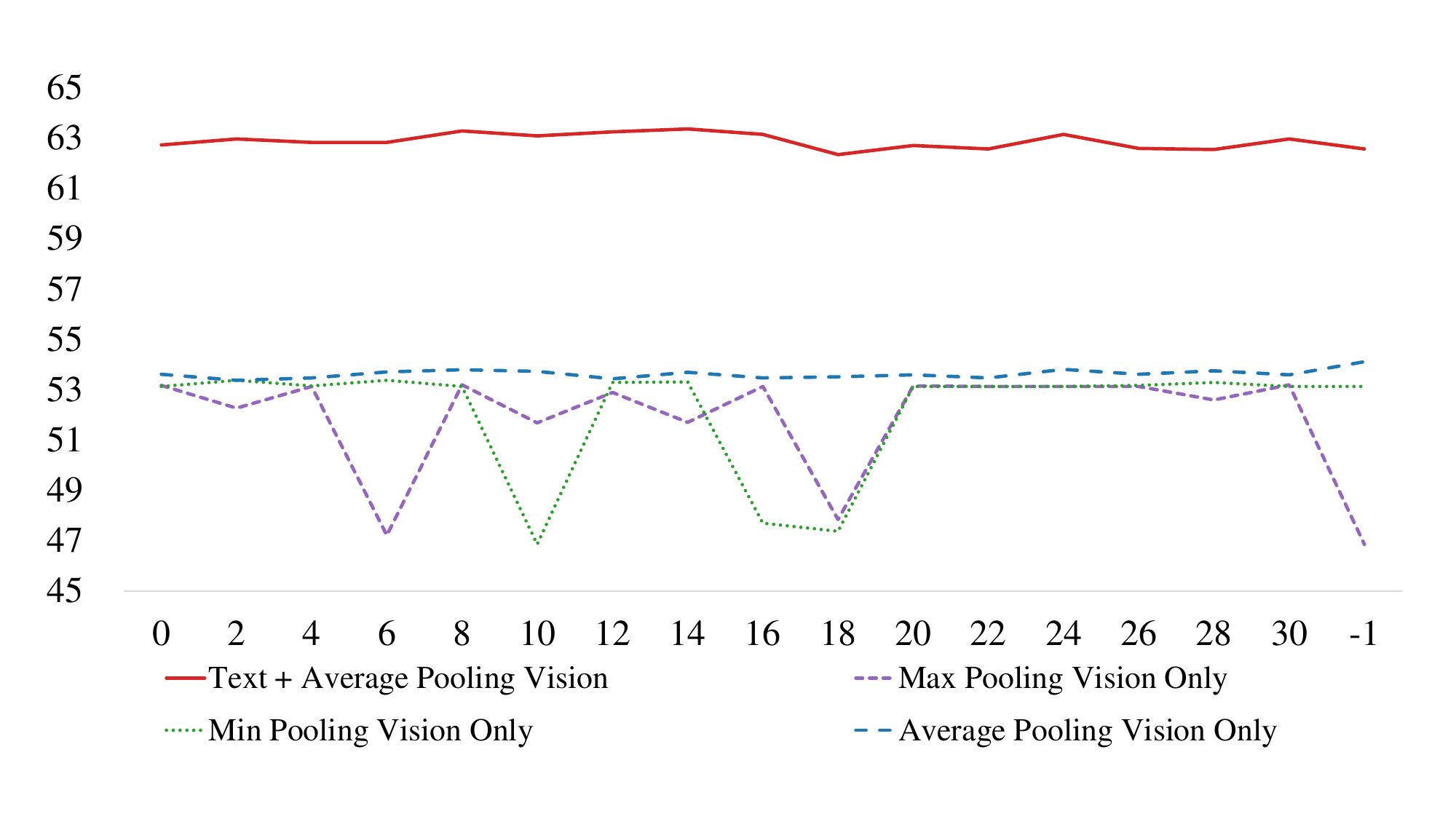}
  \caption{Layer-wise Comparison of Classifier Accuracy with Textual and Visual Features on OK-VQA Dataset}
  \label{okvqa_classifer}
\end{figure}

Additionally, we conduct a layer-wise comparative study using the Idefics2-8B model to compare classifiers that rely solely on textual features or visual features with those that jointly incorporate textual and visual features.
As shown in Fig.~\ref{classifer} and Fig.~\ref{okvqa_classifer}, the vertical axis represents the classifier accuracy, while the horizontal axis corresponds to different network layers.
The combined use of textual and visual features consistently leads to improved accuracy across datasets, and both modalities are indispensable across all layers. These results further validate the effectiveness of our approach.

\subsection{RIR Strategy Comparison}

To further examine the impact of retrieval trigger policies, we conduct a comparative study between \textit{RIR-Optimistic} and \textit{RIR-Pessimistic} strategies.
Experiments are performed with Idefics2-8B and Idefics3-8B on three datasets: InfoSeek, OK-VQA, and E-VQA.
For each dataset, we report the response accuracy under both strategies while keeping all other settings fixed.

\begin{figure}[h]
    \centering
    \includegraphics[width=\linewidth, trim=0 100 0 40, clip]{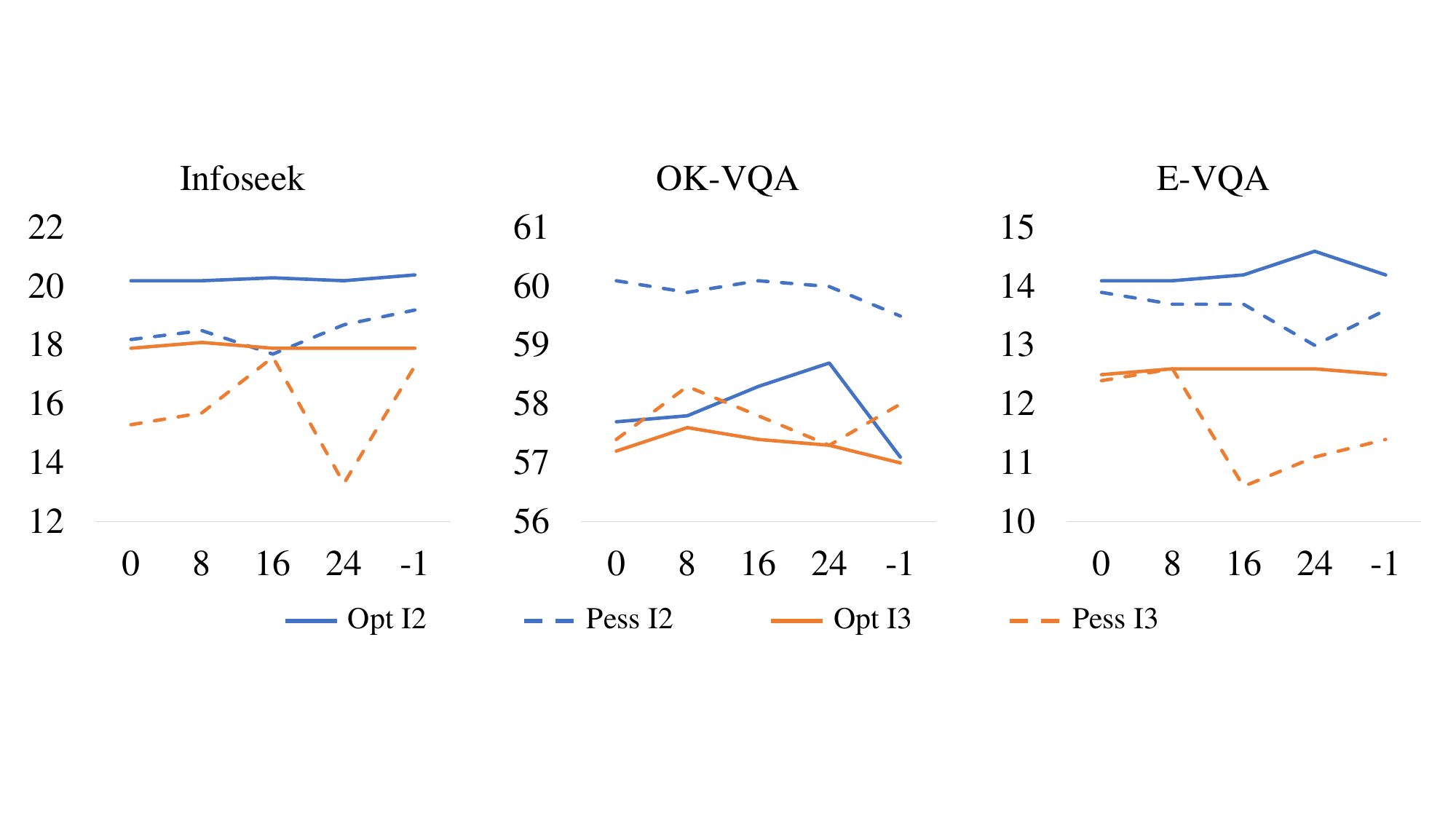}
    \caption{Performance Comparison between RIR-Optimistic and RIR-Pessimistic Strategies}
    \label{opt-vs-pess}
\end{figure}

As shown in Fig.~\ref{opt-vs-pess}, a clear dataset-dependent preference emerges between the two retrieval strategies, where the vertical axis denotes the response accuracy and the horizontal axis corresponds to the layer index.
On OK-VQA, the RIR-Pessimistic Strategy consistently outperforms the RIR-Optimistic Strategy, whereas on InfoSeek and E-VQA the opposite trend is observed.

This divergence can be attributed to the differing roles that external visual information plays across datasets. OK-VQA primarily focuses on common-sense reasoning and world knowledge, where the answer often depends less on fine-grained visual cues and more on abstract or factual understanding. In such cases, reverse image retrieval is prone to introducing visually similar yet semantically irrelevant evidence, making a pessimistic, retrieval-averse strategy more robust. By contrast, InfoSeek and E-VQA emphasize instance-level recognition and encyclopedic knowledge, where additional visual context retrieved from external sources can provide complementary cues that help disambiguate entities or attributes, thereby improving answer accuracy.
Consistent trends across the backbone indicate that strategy preference is driven by dataset characteristics, highlighting the need for adaptive retrieval policies.




\section{Conclusion}
\label{sec: Conclusion}

In this paper, we propose MMA-RAG, a multimodal adaptive retrieval-augmented generation framework that regulates external retrieval based on internal visual and textual representations of multimodal large language models.
By predicting the utility of reverse image retrieval, MMA-RAG selectively incorporates external visual evidence only when it is likely to improve response correctness, thereby mitigating harmful retrieval in visual question answering tasks.
A layer-wise analysis of multimodal internal representations reveals that the evolution of visual and textual confidence signals provides reliable cues to detect misleading evidence, motivating the use of an internal representation-based retrieval utility classifier.
Extensive experiments on multiple knowledge-intensive VQA benchmarks with diverse vision--language backbone models demonstrate that MMA-RAG improves both accuracy and inference robustness over standard retrieval-based approaches and existing baselines.

\section{References}

\bibliographystyle{IEEEtran}
\bibliography{reference}

\end{document}